\documentclass{article}

\usepackage{microtype}
\usepackage{graphicx}
\usepackage{booktabs} 

\usepackage{hyperref}

\usepackage[accepted]{latinx-icml-2021}

\icmltitlerunning{Vehicle-counting with Automatic Region-of-Interest  and Driving-Trajectory Detection}

\begin{document}

\bibliographystyle{icml2021}

\twocolumn[
    \icmltitle{Vehicle-counting with Automatic Region-of-Interest \\
        and Driving-Trajectory detection}





    \begin{icmlauthorlist}
        \icmlauthor{Malolan Vasu}{laf}
        \icmlauthor{Nelson Abreu}{uasd}
        \icmlauthor{Raysa V\'asquez}{uasd}
        \icmlauthor{Christian L\'opez}{laf}
    \end{icmlauthorlist}

    \icmlaffiliation{laf}{Lafayette College, Easton, PA, USA}
    \icmlaffiliation{uasd}{Universidad Autónoma de Santo Domingo, Dominican Republic}
    \icmlcorrespondingauthor{Christian Lopez}{lopezbec@lafayette.edu}

    \icmlkeywords{Machine Learning, Computer Vision, Vehicle Tracking, Vehicle Counting,
        Region of Interest, Representative Trajectory, Reporesentative Path, Track Clustering}

    \vskip 0.3in
]



\printAffiliationsAndNotice{}  

\begin{abstract}
    Vehicle counting systems can help with vehicle analysis and traffic incident detection. Unfortunately, most existing methods require some level of human input to identify the Region of interest (ROI), movements of interest, or to establish a reference point or line to count vehicles from traffic cameras. This work introduces a method to count vehicles from traffic videos that automatically identifies the ROI for the camera, as well as the driving trajectories of the vehicles. This makes the method feasible to use with Pan-Tilt-Zoom cameras, which are frequently used in developing countries. Preliminary results indicate that the proposed method achieves an average intersection over the union of 57.05\% for the ROI and a mean absolute error of just 17.44\% at counting vehicles of the traffic video cameras tested.
\end{abstract}

\section{Introduction}
In recent years, with the growing popularity of computer vision, and the increased performance of computational systems \cite{feng}, more complex algorithms can be run at record speeds. Following that growth and the importance of traffic management due to population growth, even more in developing countries, there has been increased  interest in road traffic surveillance and monitoring sytems \cite{AICity}. Vehicle counting is an integral part of susch systems and methods  have been proposed that achive remarkable accuracy \cite{wang, yu}.

In the AI City Challenge Workshop at CVPR (henceforth AICity), many methods for vehicle counting are presented each year \cite{AICity}. Most follow a similar three-step strategy of vehicle detection, vehicle tracking, and movement assignment from trajectory modeling and classification. These methods leverage pre-trained object detections and multi-object tracker models. In recent years, object detection has made significant progress, and models such as YOLOv4 \cite{bochkovskiy2020yolov4} provide remarkable runtime efficiency and accuracy. Multi-object tracking methods, especially those that rely on a backbone of an object detector such as DeepSORT \cite{deepSORT}, have likewise matured in performance and are frequently used for vehicle counting methods \cite{AICity}. With regards to the movement assignmet of vehicles, most of the existing methods manually define the ROI and/or Movements of Interest (MOI) as single zones, pair of entering/exit zones, or lines \cite{yu,Abdelhalim,survey,AICity}. Unfortunately, this approach limits the use of these vehicle counting methods to stationary cameras where the MOI and ROI are known a priori and do not change.

\begin{figure*}[ht!]
    \begin{center}
        \centerline{\includegraphics[width=0.8\textwidth]{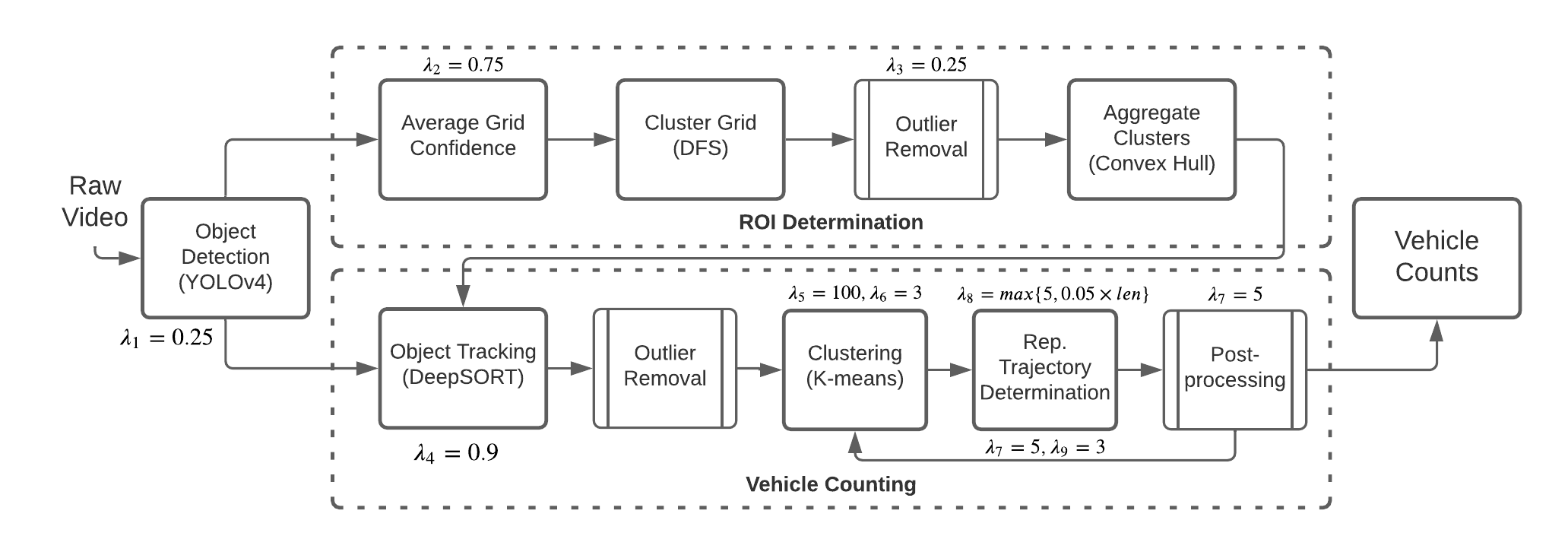}}
        \vskip -0.1in
        \caption{The two-module framework of our system.}
        \label{framework}
    \end{center}
    \vskip -0.2in
\end{figure*}

Recently, \cite{Youssef} proposed the use of a Region-based CNN and Feature Pyramid Networks to count vehicles and automatically determines the ROI from aerial video footage. While they showed promising results at counting vehicles and detecting the ROI, they were not able to detect unique MOI since they focused on counting all vehicles in footage without specifying how many vehicles were on a given driving trajectory. By leveraging existing computer vision models and techniques, this work introduces a system for vehicle counting that automatically identifies the ROI and MOI from a traffic video footage and  the vehicle count of all the driving trajectories shown in the video. The method does not require any additional user input besides the video camera footage (i.e., no ROI, MOI, or points/lines of reference). This method could find wider uses among existing traffic cameras and even work on Pan-Tilt-Zoom cameras, which are frequently used in developing countries. Hence, traffic management technology that rely on vehicle counts could benefit from this work.

\section{Method}

The proposed method utilizes two distinct modules that run in a sequential unsupervised manner. An overview is illustrated in Figure \ref{framework}. An input video is fed into the Object Detector, and its detections are used for both the first-module, which determines the ROI, and the second-module, which determines vehicle counts in driving-trajectory.

\subsection{Object Detection}
Prior research suggested that YOLOv4 provides the optimal trade-offs between speed and accuracy with a $43.5\%
    AP\ (65.7\% AP_{50}$) on the MS COCO dataset at a real-time speed of $\sim 65$ FPS on Tesla V100 \cite{bochkovskiy2020yolov4}. In this work, both cars and trucks detections found using the default YOLOv4 confidence threshold of $\lambda_1 = 0.25$ were combined into just a vehicle detection class. YOLOv4 outputs bounding boxes along with confidence values for each object detected in the camera frame used in the next step.

\subsection{ROI Determination Module}
The main assumption this works makes is that the camera is positioned such that it has maximal clarity of the ROI the user of the camera is interested in. It follows that objects detected in the ROI would have high detector confidence values compared to objects farther away from or not in the ROI.

\subsubsection{Average Grid Confidence}
Once all objects in a sequence of video frames are detected, the footage area is divided into square grids of size $max((median(obj\_widths), median(obj\_heights)),$ where object widths and heights refer to the bounding box width and height for all detections throughout the video. For every grid in the image, the average detector confidence value for the center of detection lying in the given grid is calculated. All grids above a threshold value $\lambda_2 = 0.75$, are selected for the next step. Detections are averaged out over a whole grid due to increased computational efficiency achieved by storing bounding box centers and then approximating their area instead of computing average confidence values at each pixel.

\subsubsection{Clustering of Grids}
Subsequently, all grids of average confidence exceeding $\lambda_2$ are clustered using a simple Depth-First search (DFS) where each grid is considered to be connected to the 9 grids around it. This results in many connected components of grids within the video footage area. Each cluster represents a broad region with a sizable flow of traffic where the detector has high confidence.

\subsubsection{Outlier Removal}
Sometimes, a few vehicles may be detected at the edges of the camera frame without much movement (e.g. parked vehicles) but with high confidence. Clusters with area under $\lambda_3 = 0.25$ of the average cluster size are removed from future calculations.

\subsubsection{Aggregation of Clusters}
Finally, for each of the clusters, grid cells lying on their extreme edges are searched for. The four vertices of each grid cells are then used to find the convex hull that encloses all the clusters. This represents the total ROI of this position of the camera frame where the detector has sufficient accuracy and confidence. Figure \ref{counted_cam_1} show an example of the output of the ROI Determination module were the ROI is estimated based on the confidence of the objects detected. This ROI, along with the output of the object detector, are used in the subsequent Vehicle Counting module.

\subsection{Vehicle Counting Module}
This module only considers detections within the estimated ROI from the first-module. Using the object detections, the object tracker stitches together trajectories for all vehicles passing through the ROI. Trajectories are then clustered, and a representative driving trajectory is obtained for each cluster. Finally, the number of vehicles in each of those clusters is counted.

\subsubsection{Object Tracking}
To stitch together individual object detections to an object's track across multiple frames of the video, the DeepSORT \cite{deepSORT} object tracker with a YOLOv4 backbone is used with default parameters, except $max\_iou\_distance = 0.7$ was increased to $\lambda_4 = 0.9$ to minimize ID switches, which occur when the tracker incorrectly believes a vehicle has gone out of the frame and assigns the same vehicle a new ID. Once all trajectories are obtained, they are clustered.

\subsubsection{Trajectory Clustering}
In this work, the $k$-means clustering algorithm is used with $k \in \{2,\dots, 15\}.$ Any video can expect to have at least 2 lanes (i.e. MOIs) and it is assumed most intersections would not have over 15 MOIs. A silhouette-index with a Euclidean distance measure is used and the $k$ with the maximum silhouette-index is chosen. Each vehicle track is clustered using its first and last coordinates, the difference between those two, and the angle between those two  multiplied by a factor $\lambda_5 = 100$ (i.e. clustering based on position and the displacement vector ). Clusters with less than $\lambda_6 = 3$ tracks are removed, each track is deleted, and all tracks are re-clustered with the new $k$-value.

\subsubsection{Representative Trajectory Determination}

Having obtained $k$-clusters with each object's track, they are 'averaged' out to find the representative driving trajectory for the cluster. In this work, a modified version of the method introduced by \cite{traclus} for determining the representative trajectories is implemented. Specifically, a double-sweep method is used to compute two average $y$-values, one for vectors pointing in the same direction as $\vec{V}$ (the average vector for all segments of the cluster), and the other for the opposite direction. In a regular curve, this
would provide two distinct averages for when the $x$-value
intersects two distinct paths of the cluster thereby providing
the true representative path in most cases.

The modified implementation uses Quad-Trees to efficiently work with 2D coordinates. Only vectors having $x'$ such that $|x' - x| < ((\lambda_7 = 5) \times grid\_size),$ are searched for potential intersections with the sweep line. The minimum number of lines required to intersect with a single $x$-value to be considered part of the representative path is set to $\lambda_8 = max\{5,0.05 \times num\_tracks\_in\_cluster\}.$ The hyperparameter is a proportion of the number of tracks as longer videos may have a higher proportion of outliers. Paths in both directions of the representative path with less than $\lambda_9 = 3$ points are removed. For a smooth path, the minimum distance between consecutive $x$-values is set to $\gamma = grid\_size$.

\subsubsection{Post-processing and Counting}
Next, outlier tracks that jump across the camera frame are removed by comparing each track with the representative path for that trajectory. When a significant deviation from the representative path occurs, (i.e., when at least one point on the vehicle's trajectory is further than $(\lambda_7 = 5) \times grid\_size$ from any point on the representative path), the track is deleted from the cluster. After the removals, the tracks are re-clustered and representative paths are re-calculated. Finally, the number of tracks in each cluster is calculated.

\begin{figure}[t]
    \begin{center}
        \centerline{\includegraphics[width=0.83\columnwidth]{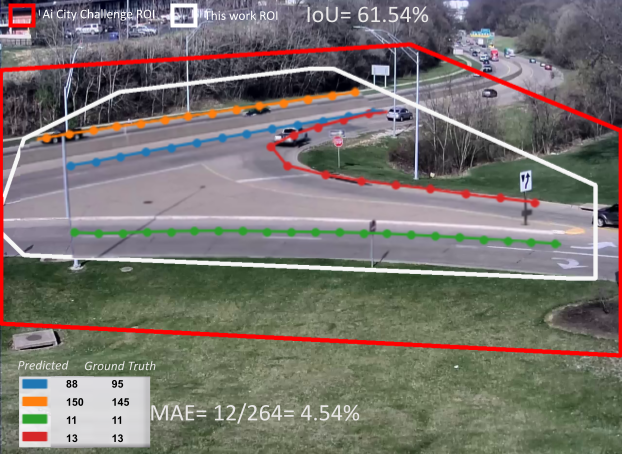}}
        \vskip -0.1in
        \caption{Final vehicle counts and rep paths for cam\_1.}
        \label{counted_cam_1}
    \end{center}
    \vskip -0.4in
\end{figure}

\section{Experimental Evaluation}
In this work, pre-trained YOLOv4 and DeepSORT models were used to perform the experiments. All experiments were ran on Google Colab notebooks with a two-thread Xeon Haswell CPU with 2.3 GHz, 12GB of RAM, and a Tesla T4 with 16GB VRAM, and are based on a subset of the AICity 2020 Track1 dataset \cite{AICity}. The hyperparameters chosen for the experiments were based on a qualitative understanding of the models and their applications, and not on any quantitative data or tunning.

\begin{table}[b]
    \vskip -0.2in
    \caption{Summary of MAE and IoU scores on the validation set.}
    \label{summary}

    \begin{center}
        \begin{small}
            \begin{sc}
                \begin{tabular}{lcccr}
                    \toprule
                    Camera Num & No.of MOIs & MAE     & IoU     \\
                    \midrule
                    1          & 4          & 4.55\%  & 61.54\% \\
                    2          & 4          & 19.83\% & 28.76\% \\
                    4          & 12         & 29.95\% & 41.25\% \\
                    5          & 12         & 26.85\% & 46.77\% \\
                    8          & 6          & 13.16\% & 55.28\% \\
                    10         & 3          & 21.26\% & 32.98\% \\
                    11         & 3          & 15.87\% & 84.32\% \\
                    14         & 2          & 4.65\% & 72.38\% \\
                    15         & 2          & 34.58\% & 66.00\% \\
                    16         & 2          & 3.70\% & 81.24\% \\
                    \bottomrule
                \end{tabular}
            \end{sc}
        \end{small}
    \end{center}
    \vskip -0.1in
\end{table}

Table \ref{summary} shows the the Mean Absolute Error (MAE) for the vehicle counting of each video tested, as well as the Intersection-over-Union (IoU) between the ROI estimated and the ground truth ROI provided by AICity. The results show that, on average, the proposed method achieved an IoU score of \textbf{57.07\%} and an MAE of \textbf{17.44\%}. For example, Figure \ref{counted_cam_1} shows the IoU and the MAE calculation for 'cam\_1' of the dataset.

\begin{figure}[h]
    \begin{center}
        \centerline{\includegraphics[width=0.8 \columnwidth]{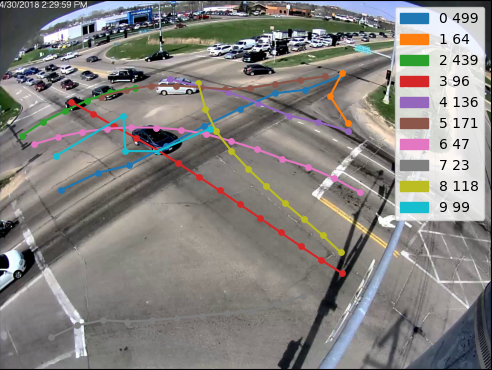}}
        \caption{Rep paths for cam\_5. True $k = 12$, predicted $k = 10$.}
        \label{counted_cam_5}
    \end{center}
    \vskip -0.3in
\end{figure}

The low IoU scores could be explained by the assumption made in this work for estimating the ROI which is not necessarily followed by the ROI provided by AICity.  For example, Fig \ref{counted_cam_1} compares the ROI estimated in this work and that in AICity. In some cases the ROI provided by AICity doesn’t have enough footage for certain paths or cover non-driving areas (e.g., grass), hence resulting in differing outcomes.

Moreover, most counting false negatives occur due to missed object detections and false positives due to ID switches, which are a result of the under-performance of DeepSORT and may require advanced methods  to alleviate, as presented in AICity\cite{wang, yu}. A preliminary overview suggests that most of counting errors stem from miscalculated $k$ for the $k$-means clustering (see Figure \ref{counted_cam_5}).

The method currently runs at approximately $11$ FPS (object detection at approx. 20 FPS, object tracking at 28 FPS, ROI at 581 FPS, clustering at 822 FPS) on the current system configuration (note that object detection can reach speeds of $65$ FPS \cite{bochkovskiy2020yolov4} on ideal system configurations), but would see much higher speeds with better processing power.

\section{Conclusion and Further Work}
This paper proposes a method of vehicle counting that is capable of automatically estimating the Region-of-Interest and driving-trajectories of vehicles without any information about the traffic footage. While the results are promising, given that the method achieved an average MAE of 17.44\% without using any a priori information about the footage, several limitations and areas of improvement still exist. As shown in Table \ref{summary}, the method's performance has high variability across videos, possibly indicating poor generalizability. Analysing the impact of the different hyperparameters used, and tuning the hyperparameters on the validation set could greatly improve generalizability and performance. Clustering algorithms specifically suited for working on vehicle trajectories could also prove useful. Future work could also explore the possibility of allowing the model to work in real-time and update its ROI, MOIs, and vehicle count as new video frames are provided.


\bibliography{traffic}

\begin{thebibliography}{10}
\providecommand{\natexlab}[1]{#1}
\providecommand{\url}[1]{\texttt{#1}}
\expandafter\ifx\csname urlstyle\endcsname\relax
  \providecommand{\doi}[1]{doi: #1}\else
  \providecommand{\doi}{doi: \begingroup \urlstyle{rm}\Url}\fi

\bibitem[Abdelhalim \& Abbas(2020)Abdelhalim and Abbas]{Abdelhalim}
Abdelhalim, A. and Abbas, M.
\newblock Towards real-time traffic movement count and trajectory
  reconstruction using virtual traffic lanes.
\newblock In \emph{Proceedings of the IEEE CVPR Workshops}, June 2020.

\bibitem[Bochkovskiy et~al.(2020)Bochkovskiy, Wang, and
  Liao]{bochkovskiy2020yolov4}
Bochkovskiy, A., Wang, C.-Y., and Liao, H.-Y.~M.
\newblock Yolov4: Optimal speed and accuracy of object detection.
\newblock In \emph{arXiv preprint arXiv:2004.10934}, 2020.

\bibitem[Feng et~al.(2019)Feng, Jiang, Yang, Du, and Li]{feng}
Feng, X., Jiang, Y., Yang, X., Du, M., and Li, X.
\newblock Computer vision algorithms and hardware implementations: A survey.
\newblock \emph{Integration}, 69:\penalty0 309--320, 2019.


\bibitem[La et~al.(2020)La, Ha, Nguyen, and Nguyen]{survey}
La, H.-P., Ha, M.-T., Nguyen, H.-L., and Nguyen, M.-T.
\newblock Vehicle counting: Survey and experiments.
\newblock \emph{7th NAFOSTED NICS}, pp.\  350--355, 2020.

\bibitem[Lee \& Han(2007)Lee and Han]{traclus}
Lee, J. and Han, J.
\newblock Trajectory clustering: A partition-and-group framework.
\newblock In \emph{SIGMOD}, pp.\  593--604, 2007.

\bibitem[Naphade et~al.(2020)Naphade, Wang, Anastasiu, Tang, Chang, Yang,
  Zheng, Sharma, Chellappa, and Chakraborty]{AICity}
Naphade, M., Wang, S., Anastasiu, D., Tang, Z., Chang, M.-C., Yang, X., Zheng,
  L., Sharma, A., Chellappa, R., and Chakraborty, P.
\newblock The 4th ai city challenge.
\newblock In \emph{Proceedings of the IEEE CVPR Workshops}, 2020.

\bibitem[Wang et~al.(2020)Wang, Bai, Xing, Zhong, Zhouqinqin, Meng, Xu, Song,
  Xu, Hu, and Chai]{wang}
Wang, Z., Bai, B., Xing, T., Zhong, B., Zhouqinqin, Z., Meng, Y., Xu, B., Song,
  Z., Xu, P., Hu, R., and Chai, H.
\newblock Robust and fast vehicle turn-counts at intersections via an
  integrated solution from detection, tracking and trajectory modeling.
\newblock In \emph{Proceedings of the IEEE CVPR Workshops}, 2020.

\bibitem[Wojke et~al.(2017)Wojke, Bewley, and Paulus]{deepSORT}
Wojke, N., Bewley, A., and Paulus, D.
\newblock Simple online and realtime tracking with a deep association metric.
\newblock In \emph{ Proceedings IEEE ICIP}, pp.\  3645--3649, 2017.

\bibitem[Youssef \& Elshenawy(2021)Youssef and Elshenawy]{Youssef}
Youssef, Y. and Elshenawy, M.
\newblock Automatic vehicle counting and tracking in aerial video feeds using
  cascade region-based convolutional neural networks and feature pyramid
  networks.
\newblock \emph{Transportation Research Record}, 2021.

\bibitem[Yu et~al.(2020)Yu, Feng, Qian, Liu, and Hauptmann]{yu}
Yu, L., Feng, Q., Qian, Y., Liu, W., and Hauptmann, A.~G.
\newblock Zero-{VIRUS}: Zero-shot {VehIcle} route understanding system for
  intelligent transportation.
\newblock In \emph{Proceedings of the IEEE CVPR Workshops}, 2020.

\end{thebibliography}

\onecolumn
\section*{ACKNOWLEDGMENT}
This research was funded by the National Fund for Innovation and Scientific and Technological Development (FONDOCYT for its acronym in
Spanish) from the Ministry of Higher Education, Science and Technology of the Dominican Republic (FONDOCYT 2018-19-3A1-107). Any opinions, findings, or conclusions found in this paper are those of the authors and do not necessarily reflect the views of the sponsors. The authors would also like to thank the hard work of the Research Assistants Orlin Cury and Manuel Garcia.

\section*{LatinX Contribution}
Three of the co-authors identify themselves as LatinX. They are the Co-PIs (second and third authors ) and the PI (last author) of a grant from the Dominican Republic government.  The second and third authors helped with the Experimental Evaluation section as well as with the data management. The first and last authors worked on the proposed method and the manuscript.

\end{document}